\documentclass[letterpaper, 10 pt, conference]{ieeeconf}  

\IEEEoverridecommandlockouts                              

\usepackage{graphics}
\usepackage{graphicx}
\usepackage{pdfpages}
\usepackage{color}
\usepackage{amsmath,amssymb}
\usepackage{mathtools}
\usepackage{color, soul}
\usepackage{multirow}
\numberwithin{equation}{section}
\renewcommand\thesection{\arabic{section}}
\usepackage{caption}
\usepackage{multicol}
\usepackage{xcolor}

\usepackage[font=small,labelfont=bf,margin=\parindent,tableposition=top]{caption}
\usepackage{subcaption}
\usepackage{titlesec}
\usepackage[colorlinks=true, allcolors=black]{hyperref}

\titlespacing*{\section}{0pt}{0.5\baselineskip}{0.2\baselineskip}
\titlespacing*{\subsection}{0pt}{0.35\baselineskip}{0.35\baselineskip}

\usepackage{algorithm,algcompatible,amsmath}
\algnewcommand\INPUT{\item[\textbf{Input:}]}%
\algnewcommand\OUTPUT{\item[\textbf{Output:}]}%
\algnewcommand\algorithmicreturn{\textbf{return}}
\algnewcommand\RETURN{\State \algorithmicreturn}

\usepackage{widetext}
\usepackage{cite}
\makeatletter
\let\NAT@parse\undefined
\makeatother

\title{\LARGE \bf
Optimizing Fuel-Constrained UAV-UGV Routes for Large Scale Coverage: Bilevel Planning in Heterogeneous Multi-Agent Systems

}

\author{Md Safwan Mondal$^{1}$, Subramanian Ramasamy$^{1}$, Pranav Bhounsule$^{1}$
\thanks{*This work was supported by ARO grant
W911NF-14-S-003.}
\thanks{$^{1}$Md Safwan Mondal, Subramanian Ramasamy and 
Pranav A. Bhounsule are with the Department of Mechanical
and Industrial Engineering, University of Illinois Chicago, IL,
60607 USA.
        {\tt\small mmonda4@uic.edu, sramas21@uic.edu, pranav@uic.edu}}%
\thanks{
        {}}%
}

\begin{document}

\maketitle
\thispagestyle{empty}
\pagestyle{empty}

\begin{abstract}

Fast moving unmanned aerial vehicles (UAVs) are well suited for aerial surveillance, but are limited by their battery capacity. To increase their endurance UAVs can be refueled on slow moving unmanned ground vehicles (UGVs). The cooperative routing of UAV-UGV multi-agent system to survey vast regions within their speed and fuel constraints is a computationally challenging problem, but can be simplified with heuristics. Here we present multiple heuristics to enable feasible and sufficiently optimal solutions to the problem. Using the UAV fuel limits and the minimum set cover algorithm, the UGV refueling stops are determined. These refueling stops enable the allocation of mission points to the UAV and UGV. A standard traveling salesman formulation and a vehicle routing formulation with time windows, dropped visits, and capacity constraints is used to solve for the UGV and UAV route, respectively. Experimental validation on a small-scale testbed ({\hypersetup{urlcolor=blue}\url{http://tiny.cc/8or8vz}}) underscores the effectiveness of our multi-agent approach.
\end{abstract}
\vspace{-2mm}
\section{INTRODUCTION}
Multi-agent systems involving Unmanned Aerial Vehicles (UAVs) and Unmanned Ground Vehicles (UGVs) are increasingly finding applications in diverse fields such as surveillance, search and rescue, and transport, owing to their collaborative advantage \cite{stolfi2021uav, wu2020cooperative,liu2019cooperative,li2016hybrid}. One key obstacle in such contexts is the UAVs' limited fuel capacity, which restricts their operational duration and reach. However, effective multi-agent cooperation between UAVs and UGVs can boost mission efficiency and extend UAV coverage, facilitating sustained, long-range operations. The cooperative routing of such multi-agent systems is complex due to its combinatorial nature, making the problem computationally intensive to solve with exact methods. Thus, the use of suitable heuristics becomes crucial for quickly attaining high-quality solutions.

\subsection {Related works}

\begin{figure*}[htpb]
\centering
\begin{subfigure}[b]{0.32\textwidth}
         \centering
         \includegraphics[ scale=0.62]{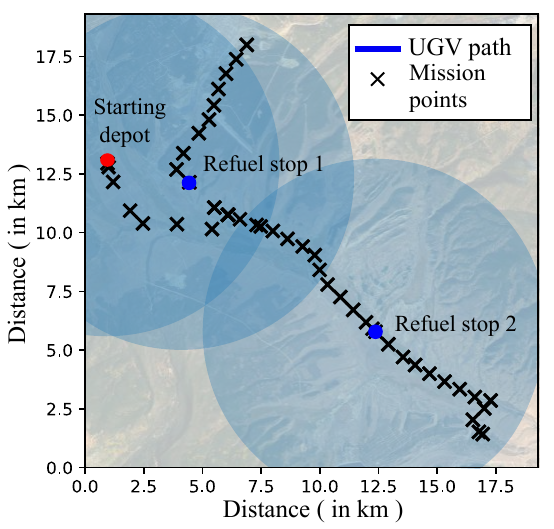}
         \caption{given mission scenario with MSC }
         \label{Entire_scenario}
         
\end{subfigure}
\hfill     
\begin{subfigure}[b]{0.32\textwidth}
         \centering
         \includegraphics[ scale=0.62]{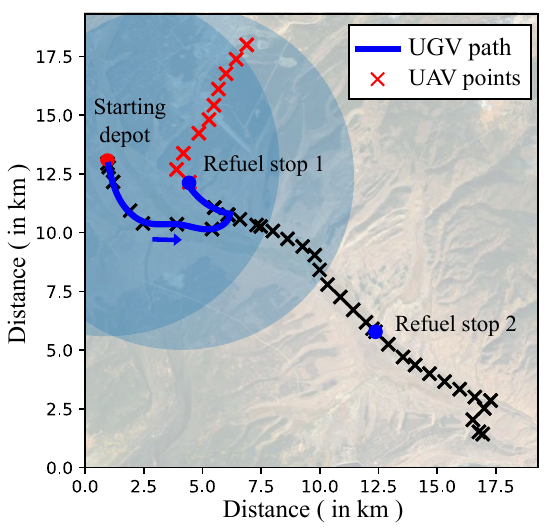}
         \caption{first subproblem}
         \label{SP1}
         
\end{subfigure}
\hfill    
\begin{subfigure}[b]{0.32\textwidth}
         \centering
         \includegraphics[ scale=0.62]{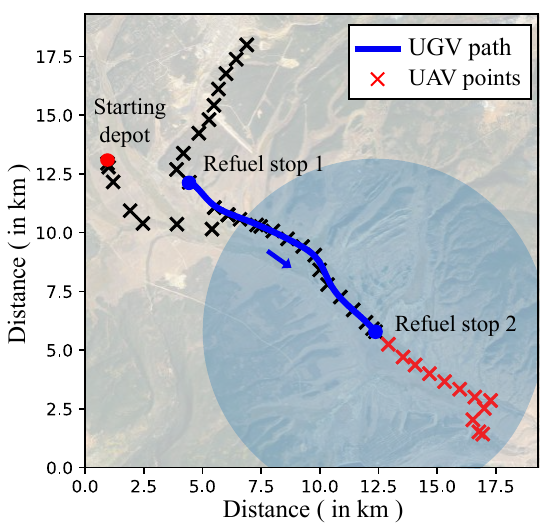}
         \caption{second subproblem}
         \label{SP2}
\end{subfigure}
\caption{Minimum set cover algorithm and task allocation technique a) Given mission scenario with minimum set cover algorithm. Blue circle indicates radial coverage of UAV. Here, 3 refuel stops (including starting depot) can cover entire mission b) First subproblem where UGV moves from starting depot to refuel stop 1 and UAV missions points within radial coverage are assigned. c) Second subproblem where UGV moves from refuel stop 1 to refuel stop 2 and UAV missions points within radial coverage are assigned. }
\vspace{-15pt}
\label{Task allocation }
\end{figure*}

Fuel constrained UAV routing has drawn significant research attention, with numerous studies focusing on routes involving multiple UAVs recharging at fixed depots. For instance, Levy et al. \cite{levy2014heuristics} used variable neighborhood descent (VND) and search heuristics (VNS) to discover viable solutions for large-scale scenarios. Sundar et al. \cite{sundar2016formulations} devised a mixed-integer linear programming (MILP) model solvable with readily available MILP solvers. Conversely, Maini et al.\cite{maini2015cooperation} considered a UAV-UGV system where the UGV acted as a moving recharging station for the UAV and they devised a greedy heuristic to identify UAV recharging points along the UGV's route. Manyam et al.  \cite{manyam2019cooperative} explored the cooperative routing problem of a UAV-UGV team, accounting for communication constraints. They modeled the issue as an MILP and introduced a branch-and-cut algorithm for optimal problem-solving.

Researchers extended the UAV-UGV cooperative vehicle routing problem by solving it in a tiered two-echelon approach. To solve the two-echelon cooperative routing problem, Luo et al. \cite{luo2017two} proposed a binary integer programming model with two heuristics. Liu et al. \cite{liu2020two} developed a two-stage routing framework for optimizing the main route of a truck and the associated flight routes of a drone in a parcel delivery system. They devised a hybrid heuristic that combined nearest neighbor and cost-cutting strategies for rapid solution development. In our previous works \cite{ramasamy2021cooperative,ramasamy2022coordinated}, we investigated a hierarchical bi-level optimization framework for the cooperative routing of multiple fuel-limited UAVs and a single UGV. The framework employed K-means clustering to generate UGV visit points and used the traveling salesman problem (TSP) to connect these points and create the UGV route. A vehicle routing problem was then formulated and solved for the UAV based on capacity constraints, time windows, and missed visits. We extended this work [13] to demonstrate that the quality of solutions could be significantly improved by optimizing heuristic parameters using Genetic Algorithm (GA) and Bayesian Optimization (BO) methods. However, the prior framework was scenario-specific, making generalization difficult, prompting this study to develop a more robust and generalized optimization framework suitable for practical hardware implementation.

On the experimental front, few significant works have been done to demonstrate localization and mapping of UAVs in indoor environments. Nigam et al. \cite{nigam2011control,nigam2014multiple,nigam2009control} investigated for high-level scalable control techniques for unmanned aerial vehicles (UAVs) for performing persistent surveillance in an uncertain stochastic environment in a hardware testbed. Two UAVs were used by Frew et al. \cite{frew2004flight} to demonstrate road following, obstacle avoidance, and convoy protection in a fligt testing, while Jodeh et al. \cite{jodeh2008overview} provided an overview of cooperative control algorithms of heterogeneous UAVs by the Air Force Research Laboratories (AFRL). Leahy et al. \cite{leahy2016persistent,leahy2016provably} experimentally validated their proposed method for automating persistent surveillance missions involving multiple vehicles. Automata-based techniques were used to generate collision-free motion plans and vector fields were created for use with a differential flatness-based controller, allowing vehicle flight and deployment to be fully automated according to the motion plans. They used charging platforms for the UAVs for truly persistent missions. Boeing’s Vehicle Swarm Technology Laboratory (VSTL) \cite{nigam2009control,redding2011distributed,saad2009vehicle,halaas2009control} and MIT’s RAVEN laboratory \cite{how2007multi} testbed have conducted significant UAV flight testing demonstrations in indoor lab scale setups.

  In this paper, we introduce a bi-level optimization frame-
work for solving the fuel-constrained UAV-UGV cooperative
routing problem that optimizes the operational time and fuel
consumption of both vehicles. Our proposed framework leverages a minimum set cover algorithm and task allocation method, enhancing the efficiency of the cooperative multi-agent system and addressing fuel and speed constraints. To validate our proposed algorithm, we conducted hardware testing on a laboratory testbed that provides practical insights into the real-world application and feasibility of our proposed approach.
   To this end, we present following novel contributions: 1) The overall framework uses bilevel optimization with task allocation technique for mission allocation and constrained programming based routing solvers. 2) The task allocation technique based on minimum set cover algorithm divides the entire problem into decoupled subproblems which radically simplifies the overall problem. 3) A constraint programming-based formulation for vehicle routing problem with time windows, dropped visits, and fuel constraints enables quick solutions of each subproblem. 4) Hardware validation of our work demonstrates the practical feasibility and real-world applicability of our proposed algorithms and methods.

\subsection{Problem Description}

The aim of the problem is to perform a cooperative mission involves visiting a set of designated mission points (see figure \ref{Entire_scenario}) $\mathcal{M} = \{m_0,m_1,...,m_n\}$, using either the\textbf{ UGV road-based visit $\tau^g$} or the \textbf{UAV aerial flyover $\tau^a$}. The cost of travel between any two mission points is defined as the time required to travel from one point to another, $t_{ij} = t_j - t_i$. The UGV and UAV are heterogeneous in nature, with the UAV having a higher velocity $v^a > v^g$ but a lower fuel capacity compared to the UGV $f^a < f^g$.  Unlike UAV, the UGV is restricted to travel along the road network only, and the fuel consumption rate of the UAV is a function of its velocity, $\mathcal{P}^a = f(v^a) $. The UAV can be recharged by the UGV at any refueling stop or at the starting depot, with the recharging time dependent on the fuel level of the UAV. The UGV is assumed to have an infinite fuel capacity due to its larger fuel capacity compared to the UAV. With these described assumptions, the objective is to find the quickest route, $\tau_{min} = \tau^a \cup \tau^g $ for the UAV and UGV to visit all mission points together, with the starting depot being both the starting and ending point, while ensuring that the UAV never runs out of fuel. 

To find the time-optimized route, the following goals must be achieved:

a) Identification of suitable stop locations \textbf{where} UAV will synchronize with UGV to get recharged to cover all the mission points, i.e, $\mathcal{M}_r = \{m^r_0,m^r_1,...,m^r_n\}$.

b) Determination of optimal times during the mission, \textbf{when} UAV, UGV will meet at those refuel stops i.e, $t^r_i \ \forall\ m^r_i \in \mathcal{M}_r $. 

c) Determination of the \textbf{optimal routes} $\tau^a , \tau^g $ for both the UGV and UAV based on the refueling locations $m^r_i$ and times $t^r_i$. 

\section{OPTIMIZATION FRAMEWORK}


For the UAV-UGV cooperative routing, we proposed a bilevel optimization framework. The framework is designed as a two-level hierarchical structure, where at the higher level, we determine the UGV route using the ``UGV First, UAV Second" heuristic method, which involves prioritizing the UGV route and then constructing the UAV route based on it. To ensure the feasibility of the cooperative route, it is critical to locate suitable refueling sites $\mathcal{M}_r$ along the UGV route. We employed a minimum set cover algorithm to identify the best locations for refueling. Then the inner-level UAV route was built based on the UGV route by dividing the entire scenario into subproblems what could be solved by modeling them as an energy constrained vehicle routing problem with time windows (E-VRPTW).

\subsection{Outer level: UGV route}

Maini et al. \cite{maini2015cooperation} demonstrated that in order to establish a viable cooperative route, it is necessary to ensure that at least one refueling stop is located within the UAV fuel coverage radius for each mission point. Thus to determine the minimum number of refueling stops required to cover the entire mission scenario, we can adopt the minimum set cover algorithm (MSC). This is a well-established problem that can be solved using a variety of methods, including greedy heuristics \cite{maini2015cooperation}. However, in this study, we proposed a constraint programming formulation for minimum set cover algorithm. For the same scenario, we employed both greedy method and constraint programming approach individually to solve the minimum set cover problem where constraint programming method outperformed greedy heuristics.  
\subsubsection{Greedy heuristics method}

Using a greedy heuristic approach can help to reduce the complexity of the minimum set cover problem. Beginning with a set of mission points $\mathcal{M}$\, that require coverage and considering the UAV's fuel capacity $f^a$ as key inputs, the objective is to ascertain the smallest possible subset $\mathcal{M}_r $ of  $\mathcal{M}$ that can act as refueling stops, to ensure coverage of the entire mission scenario. The greedy algorithm selects the initial depot point $m_0$ as the first refueling stop $m^r_0$ and then sequentially adds the mission points that cover the greatest number of other uncovered mission points to the refueling stop set $\mathcal{M}_r$ until all points are covered.

Although the greedy heuristic can produce an optimal result for a minimum set cover problem quickly, there is possibility of multiple optimal solutions for a given scenario. Since we are implementing a bilevel optimization framework, it is essential to consider all the other optimal solutions of the outer level algorithm. As it is not possible to acquire all optimal solutions using the greedy heuristic, we employed the constraint programming method. This approach can rapidly generate multiple optimal results, if present. Also, the greedy method may result into locally optimal solution which can be overcome through an alternate constraint programming formulation of the minimum set cover problem.

\begin{algorithm}[htbp]
    \caption{Greedy Minimum Set Cover Algorithm}
    \begin{algorithmic}[1]
        \INPUT Mission scenario points $\mathcal{M}$, UAV fuel limit $f^a$, starting depot $m_0$;
        \OUTPUT Refueling stops $\mathcal{M}_r$;
        \State Initialize $\mathcal{M}_r = \{m^r_0= m_0\}$, Targets = $\mathcal{T} = \mathcal{M}$;
        \State $C_0 = Covered(m^r_0) = \{m_i: m_i \in \mathcal{T} \text{ and } \|m_i - m^r_0\| < 0.5f^a \}$;
        \State $\mathcal{T} = \mathcal{T} \setminus C_0$;
        \WHILE {$\mathcal{T} \neq \emptyset$}
        \State $m^r_{i_{\text{max}}} = \arg\max\ Covered(m_i) $;
        \State $\mathcal{M}_r = \mathcal{M}_r \cup \{m^r_{i_{\text{max}}}\}$;
        \State  $C_{\text{max}} = \{m_i: m_i \in \mathcal{T} \text{ and } \|m_i - m^r_{i_{\text{max}}}\| < 0.5f^a \}$;
        \State $\mathcal{T} = \mathcal{T} \setminus C_{\text{max}}$;
        \ENDWHILE
    \end{algorithmic}
    \label{algorithm:MSC}
\end{algorithm}

\subsubsection{Constraint programming method}

Determining the minimum number of refueling stops $\mathcal{M}_r$ needed to cover the entire mission scenario $\mathcal{M}$ can be modeled using linear integer programming and solved via constraint programming method (CP method). With the binary decision variables in Eq. \ref{cp_cns3}, $x_j$ signifying if a mission point is chosen as a refueling stop, and $y_{ij}$ indicating whether a mission point $m_i$ is allocated to a refueling stop $m^r_j$,  the objective function in Eq. \ref{cp_obj} minimizes the number of refueling stops. The constraint in Eq. \ref{cp_cns} ensures that each mission point $m_i$ has at least one refueling stop $m^r_j$ assigned to it. The constraint in Eq. \ref{cp_cns1} guarantees that a mission point  $m_i$ can be allocated to a refueling stop $m^r_j$ only if that refueling stop is selected. The constraint in Eq. \ref{cp_cns2} ensures a mission point $m_i$ is only assigned to a refueling stop $m^r_j$ within the UAV's fuel coverage radius, facilitating a round trip for the UAV from the refueling stop. 

\begin{equation}
\text{Objective:     }  \min \sum_{m^r_j \in \mathcal{M}_r}x_j 
\label{cp_obj}
\end{equation}

\text{Subject to,} 
\begin{equation}
 \sum_{m^r_j \in \mathcal{M}_r} y_{ij} \geq 1, \; \forall\ m_i \in \mathcal{M}
\label{cp_cns}
\end{equation}
\begin{equation}
y_{ij} \leq x_j, \; \forall\ m_i \in \mathcal{M}\ \text{and} \ \forall\ m^r_j \in \mathcal{M}_r 
\label{cp_cns1}
\end{equation}
\begin{equation}
y_{ij} = 0, \; \text{if} \; d_{ij} > 0.5f^a, \; \forall\ m_i \in \mathcal{M}\ \text{and} \ \forall\ m^r_j \in \mathcal{M}_r
\label{cp_cns2}
\end{equation}
\begin{equation}
y_{ij}, x_j \in \{0,1\} 
\label{cp_cns3}
\end{equation}

We utilized Google's OR-Tools\texttrademark \  CP-SAT  solver \cite{ORtools} to solve this linear integer formulation, and it can record multiple optimal solutions if they exist. After identifying the refueling stop locations, a UGV route can be created by connecting these stops on the road network. We can solve a simple travelling salesman problem (TSP) considering the refueling stops to determine an optimal UGV route. Once the optimal UGV route is established, we can proceed to the inner loop UAV routing. 

\subsection{Inner level: UAV route}

At the inner level of our framework, we employed a task allocation technique to divide the entire mission scenario into independent subproblems which were solved individually as an energy constrained vehicle routing problems with time windows (E-VRPTW).

\subsubsection{Task allocation technique}

Given the scenario and the obtained UGV route from outer loop MSC algorithm,
we can divide the entire problem into $ n - 1 $  number of subproblems ($ n $ = number of
refuel stops with starting depot) with an assumption that UGV travels only between
two refuel stops in each subproblem. Before the subproblem division, each mission point is assigned to its nearest refuel stop (including starting depot) that covers it. In the subproblems, the first refuel stop is the origin node and the second refuel stop is the destination node of UGV route. The sub-problems are decoupled from each other by allocating separate mission points in them. The UAV mission points covered by destination refuel stop under 
each subproblem are allocated to that subproblem. Only, for the first subproblem the
mission points covered by both origin and destination node should be allocated to it.

Figure \ref{Task allocation } demonstrate the process of subproblem division and mission allocation.
Figure \ref{Entire_scenario} shows refuel stop locations obtained from outer level MSC algorithm for a given
scenario. The first subproblem (figure \ref{SP1}) is created by taking the starting depot as
the origin node and the refuel stop 1 as the destination node. The UAV mission
points covered by origin node (starting depot) and destination node (refuel stop 1)
are assigned for subproblem 1. Similarly, the second subproblem (figure \ref{SP2}) is created
by taking the refuel stop 1 as origin node and refuel stop 2 as destination node and
the mission points covered by the destination node (refuel stop 2) are assigned for
this subproblem. Once we get an independent set of subproblems through task allocation, we
try to solve each subproblem by modeling it as energy constrained vehicle routing
problem with time windows (E-VRPTW).

\subsubsection{E-VRPTW model}

The formulation of the E-VRPTW can be described with the graph theory. Consider an
undirected graph $G = (V, E)$ where $V$ is the set of vertices $V = \{S,0,1,2,...D\}$ and
$E$ is the set of edges between the vertices $i$ and $j$ as $ E= \{(i, j) \, \| \ i, j \, \in \, V, i \neq j \}$. 
The non-negative arc cost between the vertices  $i$ and $j$ is expressed as $t_{ij}$ (traversal time) and $x_{ij}$ is a binary decision variable whose value will be 1 if a vehicle travels from $i$ to $j$, and 0 otherwise. From the starting depot $S$, the UAV will take off and meet the UGV at destination depot $D$ which also has a time window constraint because of slower speed of UGV. The objective function of the E-VRPTW problem is indicated by Eq. \ref{eq:1} which minimizes the total travel time. Formulation of the energy and time-window constraints are done in Eq. \ref{eq:2}-\ref{eq:6}. The details explanation of other generic VRP constraints are discussed in our previous work \cite{ramasamy2022heterogenous}.

\begin{equation}
\text{Objective: } \min \sum_i \sum_j t_{ij} x_{i j} \quad \forall i, j \in V  \label{eq:1}\\ 
\end{equation}

\text{Energy constraints:} 
\begin{equation}
f^a_j= f^a, \quad j \in D \label{eq:2}
\end{equation}
\begin{equation}
0 \leq f^a_j \leq f^a, \quad \forall j \in V \setminus \{S,D\} \label{eq:3}
\end{equation}
\begin{align}
    f^a_j &\leq f^a_i-\left(\mathcal{P}^a(v^a)t_{i j}x_{i j}\right) \nonumber \\
    &+ L_1\left(1-x_{i j}\right), \quad \forall i \in V, j \in   V \setminus \{S,D\} \label{eq:4}
\end{align}
\text{Time window constraints:}
\begin{equation}
t_{j,start} \leq t_j \leq t_{j,end}, \quad \forall j \in D \label{eq:5}
\end{equation}
\begin{equation}
t_j \geq t_i+\left(t_{i j} x_{i j}\right)-L_2\left(1-x_{i j}\right), \quad \forall i \in V, j \in V \label{eq:6}
\end{equation}

Eq. \ref{eq:2} says that UAV is fully recharged at the destination depot, whereas fuel level is between 0 and maximum capacity at any other vertex (Eq. \ref{eq:3}). Eq. \ref{eq:4} employs the Miller-Tucker-Zemlin (MTZ) formulation for subtour elimination while ensuring that during the removal of sub-tours UAV's fuel level is never completely drained. The time window constraint for destination node is put in Eq. \ref{eq:5} while Eq. \ref{eq:6} makes sure that the cumulative arrival
time at $j^{th}$ node is equal to the sum of cumulative time at
the node $i, t_i$ and the travel time between them $t_{ij}$. Again, we used Google OR-Tools \texttrademark \ for our heuristic implementation for solving the E-VRPTW model with constrained programming (CP). 

\section{RESULTS}

\begin{figure}[]

\centering
\begin{subfigure}[]{0.3\textwidth}
\centering
\includegraphics[ scale=0.58]{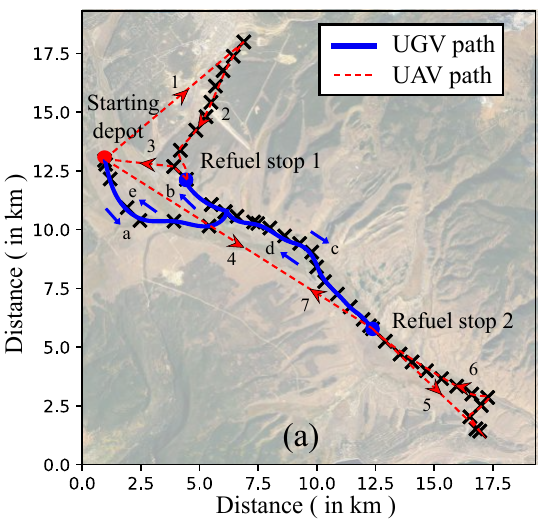}
\label{CP_method_route}
\end{subfigure}
\hfill

\begin{subfigure}[]{0.3\textwidth}

\centering
\includegraphics[ scale=0.58]{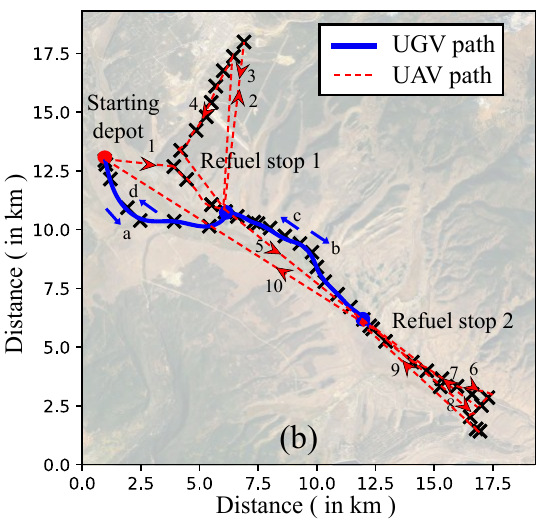}
\label{Greedy_method_route}
\end{subfigure}

\caption{UAV \& UGV trajectory obtained from bilevel optimization with Greedy and CP method at the outer loop. Numerical and alphabetical order shows the UAV and UGV motion respectively. a) CP method based trajectory b) greedy method based trajectory.  }
\label{simulation_result}
\vspace{-9pt}
\end{figure}

 Figure \ref{simulation_result} provides an illustrative example of the input and output of the problem at hand. The input, depicted in Figure \ref{simulation_result}, consists of mission points denoted by black crosses. The UAV and UGV must both initiate and terminate at the starting depot, while ensuring that all mission points are covered either by the UAV or UGV. The UAV can recharge at the depot or at designated refueling sites from the UGV. The UAV and UGV have fixed velocities of 10 m/s and 4 m/s, respectively, and their fuel capacities are 287.7 kJ and 25.01 MJ, respectively.
 
 To carry out the optimizations, we employed Python 3 and OR Tools \texttrademark \ library, which ran on a 3.7 GHz Intel Core i9 processor with 48 GB RAM on a 64-bit operating system. For the scenario, two types of cooperative routes (if different) were generated by implementing the Greedy method and the CP method at the outer loop of the suggested framework. The UGV-only route, where only a UGV completes the whole mission was also determined for the scenario. Based on the metrics of total mission completion time and total energy consumption, the impact of collaboration between the UAV and UGV on mission execution was assessed by comparing the cooperative route with the UGV only route which served as the upper limit.

 Table \ref{Impact} shows the improvement that was achieved through cooperative routing of UAV-UGV on the mission scenario. Cooperative routing is extremely energy efficient. Both CP method and greedy method at the outer loop showed positive improvement reducing total energy consumption in the mission by 36-39\%. However, for total mission time greedy method at the outer loop had a negative impact. This is due to position of refuel stops (see trajectory in figure \ref{simulation_result}b) what made the UAV to take frequent detours (6 times) for recharging at the refuel sites elongating the total mission time. However, appropriate refuel stop locations obtained CP method (see trajectory in figure \ref{simulation_result}a) helped UAV to complete its route with less recharging detour (4 times), which effectively reduced total mission time.

 Further insights about the trajectory of the cooperative route can be drawn from Table \ref{trajectory}. As discussed earlier, greedy results in longer UAV travel time which ultimately costs higher mission time. Energy consumption of UGV is low in greedy method as UAV is visiting majority of mission points compare to CP method. In sum, both CP method and greedy heuristics are capable of providing feasible cooperative route for constrained complex mission scenario; however CP method outperforms greedy method at the cost of some computational efficiency.

\begin{table}[]
\renewcommand{\arraystretch}{0.55}
\caption{Impact of the optimal solution of the cooperative routing.}
\begin{tabular}{lcclcc}
\hline \\
\multicolumn{1}{c}{Metrics} & \multicolumn{2}{c}{\begin{tabular}[c]{@{}c@{}}Cooperative   route \\ \end{tabular}}  & \begin{tabular}[c]{@{}c@{}}UGV only route \\\end{tabular} & \multicolumn{2}{c}{Improvement (\%)} \\ \cline{2-3} \cline{5-6} \\

 & \multicolumn{1}{l}{\begin{tabular}[c]{@{}l@{}}CP \\ method\end{tabular}} & \multicolumn{1}{l}{\begin{tabular}[c]{@{}l@{}}Greedy\\  method\end{tabular}} & \multicolumn{1}{l}{} & \multicolumn{1}{l}{\begin{tabular}[c]{@{}l@{}}CP \\ method\end{tabular}} & \multicolumn{1}{l}{\begin{tabular}[c]{@{}l@{}}Greedy \\ method\end{tabular}} \\ \hline \\

\begin{tabular}[c]{@{}l@{}}Time \\ consumption \\ (min.) \end{tabular} & 200 & 272 & \multicolumn{1}{c}{233} & 14.16 & -16.74 \\ \\
\begin{tabular}[c]{@{}l@{}}Energy \\ consumption \\ (MJ)\end{tabular} & 21.98 & 21.14 & \multicolumn{1}{c}{34.69} & 36.62 & 39.06 \\ \\ \hline
\end{tabular}
\label{Impact}
\vspace{-4pt}
\end{table}

\begin{table}[]
\renewcommand{\arraystretch}{0.3}
\centering
\caption{Comparison between trajectories of CP method and greedy heuristics }
\begin{tabular}{p{4cm}  p{1.6cm}  p{1.6cm}} 
 \hline \\[2ex]

 Metrics & CP method  & Greedy method \\ [2ex] 
\hline\\
 Total time (min) & 200  & 272  \\ [2ex]

Computational time (min) & 9 & 4 \\ [2ex]

 \textbf {UGV results} & & \\ [2ex]

Travel time  (minutes) & 200  & 272  \\  [2ex]
 Energy consumed (MJ) & 20.79  & 19.52  \\ [2ex]
 Mission visited & 22 & 18 \\ [2ex]

\textbf {UAV results} &  & \\ [2ex]
 
 Travel time (minutes) & 100  & 136.203  \\ [2ex]
  Energy consumed (kJ) & 1186.464  & 1618.092  \\ [2ex]
 Recharging stops on UGV & 3 & 6 \\ [2ex]
  Recharging stops on Depot & 1 & 0 \\ [2ex]
Missions visited  & 22 & 26 \\ [2ex]
\hline\\
\end{tabular}

\label{trajectory}
\vspace{-11pt}
\end{table}

\section{EXPERIMENT DESIGN}
The most stringent way of validating a framework is by hardware demonstration, but it often has limitations in terms of scope and variety. We often supplement these with simulation results to evaluate method performance and corroborate specific findings. Testing a surveillance planning framework through experiments is notably critical due to the problem's complexity. We created a small-scale lab scenario involving one UAV and UGV to test our proposed framework. To achieve full autonomy, each robot should independently locate, plan, and execute its route without external input, using strategies built upon the hardware's sensing, processing, and communication features. The development of an effective experimental system is challenging, as it involves integrating software, hardware, and communication. The individual elements of the hardware architecture are detailed separately (see figure \ref{Hardware}) as follows :

\begin{figure*}[htpb]
\centering
\begin{subfigure}[b]{0.36\textwidth}
         \centering
\includegraphics[ scale=0.28]{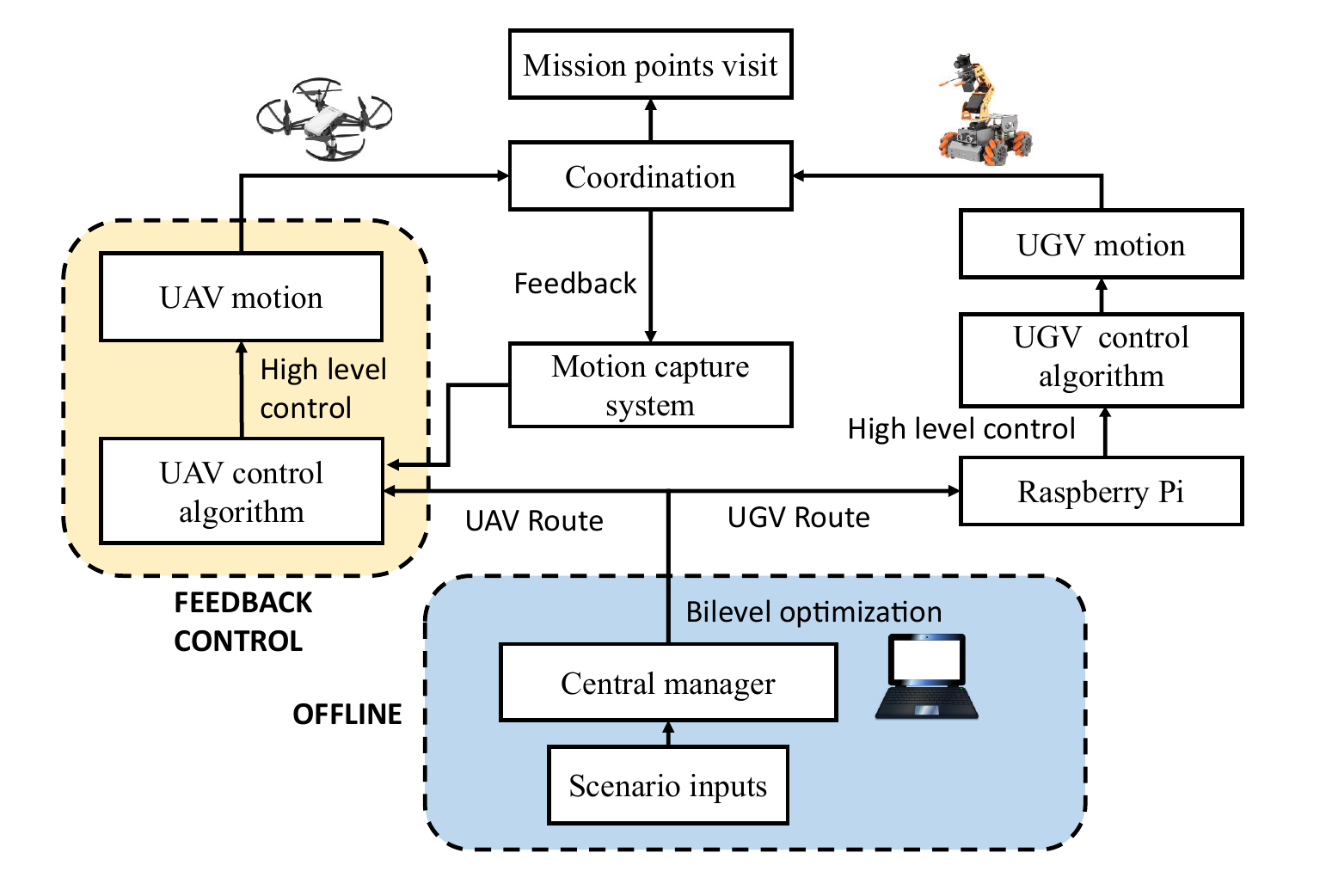}
\caption{}
\label{Hardware}
         
\end{subfigure}
\hfill     
\begin{subfigure}[b]{0.24\textwidth}
         \centering
\includegraphics[ scale=0.36]{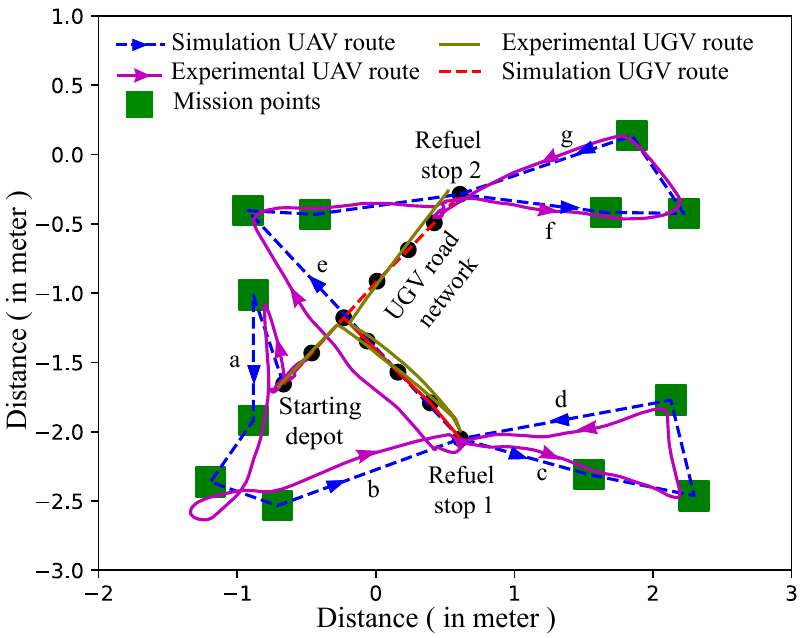}
\caption{ }
\label{Trajectory}
         
\end{subfigure}
\hfill    
\begin{subfigure}[b]{0.3\textwidth}
\centering
\includegraphics[ scale=0.35]{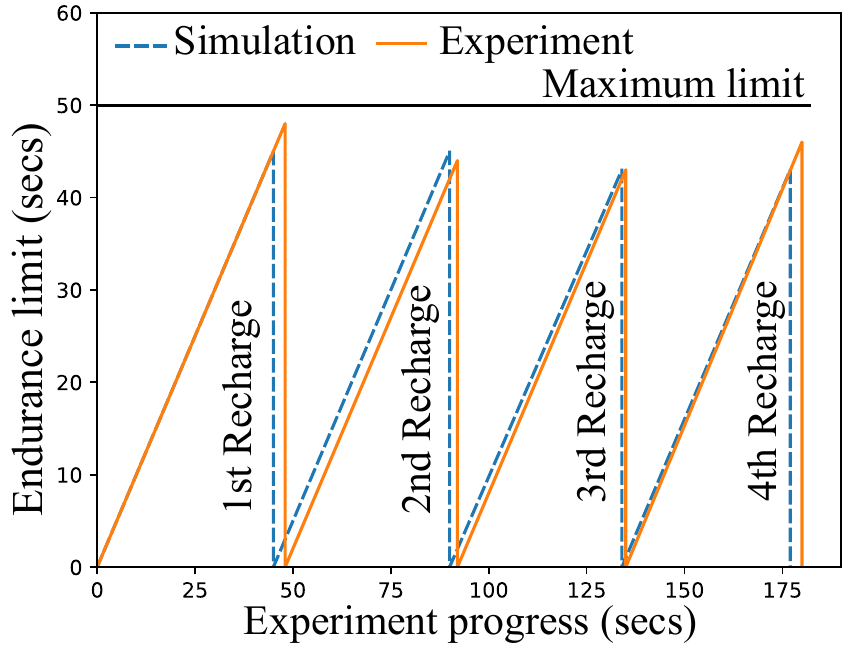}
\caption{ }
\label{Endurance_limit}
\end{subfigure}
\caption{a) Hardware architecture b) \& c) Comparison between simulation and experimental results. In the trajectory, the alphabetical order represents the direction of motion of the UAV}

\label{Task allocation }
\end{figure*}




\begin{figure*}[]
\centering
\begin{subfigure}[b]{0.495\textwidth}
         \centering
         \includegraphics[ scale=0.78]{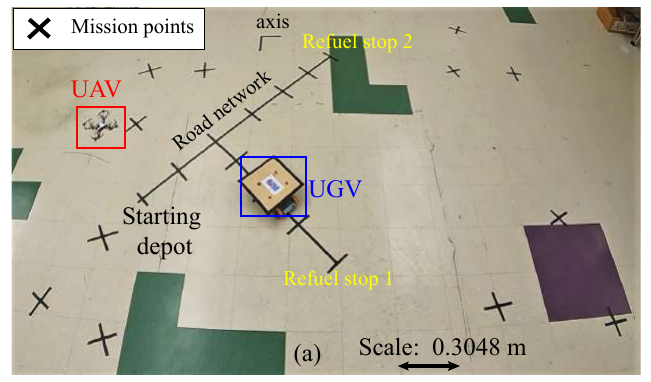}

\end{subfigure}
\hfill     
\begin{subfigure}[b]{0.495\textwidth}
         \centering
         \includegraphics[ scale=0.78]{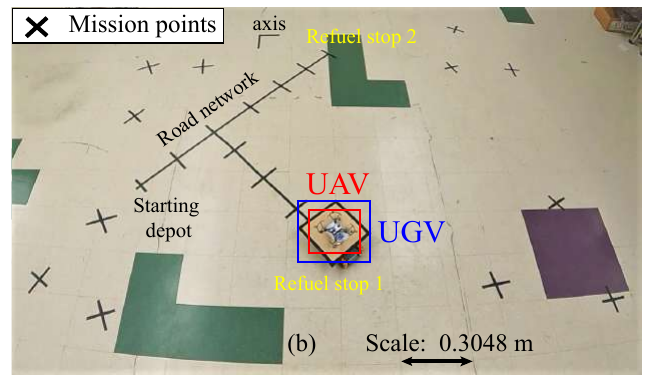}

\end{subfigure}

\caption{ Experiment instances a) UAV, UGV are moving towards their designated locations b) UAV landed on UGV for recharging. }
\label{experiment instances}
\vspace{-15pt}
\end{figure*}

1) \textbf{Hardware:} In our experiment, we selected the compact and affordable DJI Tello quadcopter as our UAV. This lightweight drone (80 g) features a closed hardware flight controller that allows basic functionality such as flight stabilization and simple trajectory execution. It can also achieve higher autonomy via an external ground-based computer using its telemetry and video feed. For our UGV, we designed a small, Raspberry Pi-controlled omnidirectional car equipped with a landing pad for recharging the UAV. The UGV's movements are managed by controlling its four wheels through the Raspberry Pi. 

2) \textbf{Control \& communication:} A wireless 2.4 GHz 802.11n WiFi connection was used to communicate with the drone. The approach makes use of the official Tello SDK 2.0. The UDP port is used to send text messages to the drone programming interface. To create the application, we used the SDK and the low-level Python library DJItelloPy. Wireless wifi communication was also established with the Raspberry Pi for controlling the UGV. 

3) \textbf{Central manager:} The final component of our system is a centralized manager. This entity employs our bi-level optimization framework to generate individual routes for the UAV and UGV, based on the input scenario. It communicates with and assigns tasks to both vehicles to initiate surveillance. The manager uses a motion capture system to continuously track their progress. The UGV, which operates on an open-loop control, halts at designated locations. Meanwhile, the UAV is guided by the central system's feedback control to ensure accurate navigation and successful recharging landings on the UGV.

4) \textbf{Experiment scenario:}  The experiments were conducted in the Robotics and Motion Lab at the University of Illinois Chicago. The lab has a designated flight area equipped with a motion capture system that serves as a reference for the position of reflective markers placed on the quadrotor and the ground vehicle. This enables real-time localization of the robots during the experiment. The positional data of the vehicles can be obtained at a rate of 100 Hz, with a latency of less than 9 $ms$. A mission scenario was created by selecting 12 different points over an area of $4 m \times 4 m $ for the UAV, a road network was designed for UGV and a fuel constraint was introduced by limiting the UAV's flight time in a single recharge (endurance limit). For this experimental setup the endurance limit was setup to be 50 seconds, the UAV and UGV speed was 0.20 $ms^{-1}$ and 0.15 $ms^{-1}$ respectively. This required the UAV to visit the UGV at regular intervals for recharging in order to complete the mission. However no real recharging took place, it was only hypothesized that UAV got recharged instantly when it landed on the UGV. The UGV road network was also designed to be challenging, with the farthest points on the mission requiring the UAV to operate near its maximum endurance limit, thus testing the robustness of the proposed framework.

\subsection{Flight test results}

Multiple trials of the experiment were carried out on the scenario. The algorithm was fed with the locations of the mission points and the road network points as input. The outer-loop of the algorithm determined the UGV traversal path with refueling spots in space and time, while the inner-loop of the framework generated the UAV route. Both the routes were provided to the individual agents from the central manager, and the agents started performing their missions. The purpose of the experiment was to verify the feasibility of the algorithm's output and to determine if the multi-agent experiment could be successfully carried out with our experimental architecture. The motion capture system was used to track the positional data of the UAV and UGV, which was processed to produce the experimental route. The figure \ref{Trajectory} shows a comparison between the simulation route and the experimental route. During the experiment, due to dynamics, the UAV drifted away in some places but successfully managed to visit the mission points and get recharged from the UGV by landing on it, because of the feedback control. The endurance limit constraint was also tested, and it was observed that the maximum flight time in a single recharge was always below the maximum limit in figure \ref{Endurance_limit}. Dynamics of the UAV played an important role in the experiment which was compensated by considering buffer time period in the modeling of take off and landing of the UAV in the simulation counterpart. Instances of the flight test can be seen in figure \ref{experiment instances}. 

\section{CONCLUSIONS}

We conclude that that a bilevel optimization framework, combined with strategically devised heuristics, offers an effective strategy for addressing multi-agent cooperative routing challenges. Our heuristics method begins by determining the UGV route through the application of the minimum set cover algorithm and traveling salesman problem, followed by establishing the UAV route based on mission allocation and a vehicle routing scheme. We found constraint programming outperform greedy heuristics to solve the minimum set cover algorithm at the expense of increased computational time. The effectiveness of our proposed multi-agent methodology is further validated by an experimental evaluation on a small testbed, which demonstrated a strong congruence between simulation and actual hardware results. 

\addtolength{\textheight}{-0cm}   





\bibliographystyle{unsrt}   
\bibliography{IROS23}



\end{document}